\documentclass[12pt,a4paper]{article}
\usepackage{times}
\usepackage{helvet}
\usepackage{courier}
\usepackage{url}
\usepackage{graphicx}
\usepackage{hyperref}
\usepackage{todonotes,bbm,bm,amsthm,amsmath,amsfonts,amssymb,mathtools,amssymb,mathrsfs,xfrac,tikz,pgf}
\newtheorem{proposition}{Proposition}
\newtheorem{example}{Example}
\newtheorem{theorem}{Theorem}
\newtheorem{transformation}{Transformation}

\newtheorem{definition}{Definition}
\newtheorem{lemma}{Lemma}
\newtheorem{myproof}{Proof of Th.}
\title{Reliable Uncertain Evidence Modeling in Bayesian Networks by Credal Networks}
\author{Sabina Marchetti\\ Sapienza University of Rome\\ Rome (Italy)\\ {\tt sabina.marchetti@uniroma1.it}
\and Alessandro Antonucci\\ IDSIA\\ Lugano (Switzerland)\\{\tt alessandro@idsia.ch}} %
\date{\today}
\begin{document}
\maketitle
\begin{abstract}
A reliable modeling of uncertain evidence in Bayesian networks based on a set-valued quantification is proposed. Both soft and virtual evidences are considered. We show that evidence propagation in this setup can be reduced to standard updating in an augmented \emph{credal} network, equivalent to a set of consistent Bayesian networks. A characterization of the computational complexity for this task is derived together with an efficient exact procedure for a subclass of instances. In the case of multiple uncertain evidences over the same variable, the proposed procedure can provide a set-valued version of the geometric approach to opinion pooling.
\end{abstract}
\section{Introduction}
Knowledge-based systems are used in AI to model relations among the variables of interest for a particular task, and provide automatic decision support by inference algorithms. This can be achieved by joint probability mass functions. When a subset of variables is observed, \emph{belief updating} is a typical inference task that propagates such (fully reliable) evidence. Whenever the observational process is unable to clearly report a single state for the observed variable, we refer to \emph{uncertain} evidence. This might take the form of a \emph{virtual} instance, described by the relative likelihoods for the possible observation of every state of a considered variable \cite{pearl88}. Also, \emph{soft} evidence \cite{valtorta2002} denotes any observational process returning a probabilistic assessment, whose propagation induces a \emph{revision} of the original model \cite{jeffrey1965}. Bayesian networks are often used to specify joint probability mass functions implementing knowledge-based systems \cite{kollerfriedman:2009}. Full, or \emph{hard} \cite{valtorta2002}, observation of a node corresponds to its \emph{instantiation} in the network, followed by belief updating. Given virtual evidence on some variable, the observational process can be modeled \emph{\`a la Pearl} in Bayesian networks: an auxiliary binary child of the variable is introduced, whose conditional mass functions are proportional to the likelihoods \cite{pearl88}. Instantiation of the auxiliary node yields propagation of virtual evidence, and standard inference algorithms for Bayesian networks can be used \cite{kollerfriedman:2009}. Something similar can be done with soft evidence, but the quantification of the auxiliary node should be based on additional inferences in the original network \cite{chan2005}. 

In the above classical setup, sharp probabilistic estimates are assumed for the parameters modeling an uncertain observation. We propose instead a generalized set-valued quantification, with interval-valued likelihoods for virtual evidence and sets of marginal mass functions for soft evidence. This offers a more robust modeling of observational processes leading to uncertain evidence. To this purpose, we extend the transformations defined for the standard case to the set-valued case. The original Bayesian network is converted into a \emph{credal network} \cite{cozman2000}, equivalent to a set of Bayesian networks consistent with the set-valued specification. We characterize the computational complexity of the credal modeling of uncertain evidence in Bayesian networks, and propose an efficient inference scheme for a special class of instances. The discussion is indeed specialized to \emph{opinion pooling} and our techniques used to generalize geometric functionals to support set-valued opinions.

\subsection{Related Work} Model revision based on uncertain evidence is a classical topic in AI. Entropy-based techniques for the absorption of uncertain evidence were proposed in the Bayesian networks literature \cite{valtorta2002,peng2010bayesian}, as well as for the pooling of convex sets of probability mass functions \cite{adamvcik2014information}. Yet, this approach was proved to fail standard postulates for revision operators in generalized settings \cite{grove1997}. Uncertain evidence absorption has been also considered in the framework of generalized knowledge representation and reasoning \cite{dubois2008three}. The discussion was specialized to evidence theory \cite{zhou2014,ma2011}, although revision based on uncertain instances with graphical models becomes more problematic and does not give a direct extension of the Bayesian networks formalism \cite{Simon2007}. Finally, credal networks have been considered in the model revision framework \cite{decampos}. Yet, these authors consider the effect of a sharp quantification of the observation in a previously specified credal network, while we consider the opposite situation of a Bayesian network for which \emph{credal} uncertain evidence is provided.

\section{Background}
\subsection{Bayesian and Credal Networks}
Let $X$ be any discrete variable. Notation $x$ and $\Omega_{X}$ is used, respectively, for a generic value and for the finite set of possible values of $X$. If $X$ is binary, we set $\Omega_X:=\{x,\neg x\}$. We denote as $P(X)$ a probability mass function (PMF) and as $K(X)$ a \emph{credal set} (CS), defined as a set of PMFs over $\Omega_X$. We remove \emph{inner} points from CSs, i.e. those which can be obtained as convex combinations of other points, and assume the CS finite after this operation. CS $K_0(X)$, whose convex hull includes all PMFs over $\Omega_X$ is called \emph{vacuous}.

Given another variable $Y$, define a collection of conditional PMFs as $P(X|Y) := \{P(X|y)\}_{y \in \Omega_Y}$. $P(X|Y)$ is called \emph{conditional probability table} (CPT). Similarly, a credal CPT (CCPT) is defined as $K(X|Y):=\{ K(X|y)\}_{y \in \Omega_Y}$. An \emph{extensive} CPT (ECPT) is a finite collection of CPTs. A CCPT can be converted into an equivalent ECPT by considering all the possible combinations from the elements of the CSs.

Given a joint variable $\bm{X}:=\{X_0,X_1, \ldots ,X_n\}$, a Bayesian network (BN) \cite{pearl88} serves as a compact way to specify a PMF over $\bm{X}$. A BN is represented by a directed acyclic graph $\mathcal{G}$, whose nodes are in one-to-one correspondence with the variables in $\bm{X}$, and a collection of CPTs $\{ P(X_i|\Pi_i) \}_{i=0}^n$, where $\Pi_i$ is the joint variable of the parents of $X_i$ according to $\mathcal{G}$. Under the \emph{Markov condition}, i.e. each variable is conditionally independent of its non-descendants non-parents given its parents, the joint PMF $P(\bm{X})$ factorizes as $P(\bm{x}):=\prod_{i=0}^n P(x_i|\pi_i)$, where the values of $x_i$ and $\pi_i$ are those consistent with $\bm{x}$, for each $\bm{x} \in \Omega_{\bm{X}} = \times_{i=0}^n \Omega_{X_i}$.

A credal network (CN) \cite{cozman2000} is a BN whose CPTs are replaced by CCPTs (or ECPTs). A CN specifies a joint CS $K(\bm{X})$, obtained by considering all the joint PMFs induced by the BNs with CPTs in the corresponding CCPTs (or ECPTs).

The typical inference task in BNs is \emph{updating}, defined as the computation of the posterior probabilities for a variable of interest given hard evidence about some other variables. Without loss of generality, let the variable of interest and the observation be, respectively, $X_0$ and $X_n=x_n$. Standard belief updating corresponds to:
\begin{equation}\label{eq:updating}
P(x_0|x_n)=\frac{\sum_{x_1,\ldots,x_{n-1}}\prod_{i=0}^n P(x_i|\pi_i)}{\sum_{x_0,x_1,\ldots,x_{n-1}}\prod_{i=0}^n P(x_i|\pi_i)}\,.
\end{equation}
Updating is NP-hard in general BNs \cite{cooper1990}, although efficient computations can be performed in polytrees \cite{pearl88} by message propagation routines \cite{kollerfriedman:2009}. 

CN updating is similarly intended as the computation of lower and upper bounds of the updated probability in Eq.~\eqref{eq:updating} with respect to $K(\bm{X})$. Notation $\underline{P}(x_0|x_n)$ ($\overline{P}(x_0|x_n)$) is used to denote lower (upper) bounds. CN updating extends BN updating and it is therefore NP-hard \cite{campos2005}. Contrary to the standard setting, inference in generic polytrees is still NP-hard \cite{maua2014a}, with the notable exception of those networks whose variables are all binary \cite{fagiuoli1998}.

\subsection{Virtual and Soft Evidence}
Eq.~\eqref{eq:updating} gives the updated beliefs about queried variable $X_0$. The underlying assumption is that $X_n$ has been the subject of a fully reliable observational process, and its actual value is known to be $x_n$. This is not always realistic. Evidence might result from a process which is unreliable and only the likelihoods for the possible values of the observed variable may be assessed (e.g., the precision and the false discovery rate for a positive medical test). \emph{Virtual evidence} (VE) \cite{pearl88} applies to such type of observation. Notation $\lambda_{X_n}:=\{\lambda_{x_n}\}_{x_n\in\Omega_{X_n}}$ identifies a VE, $\lambda_{x_n}$ being the likelihood of the observation provided $\left(X_n=x_n\right)$. Given VE, the analogous of Eq.~\eqref{eq:updating} is:

\begin{equation}\label{eq:updatingVE}
P_{\lambda_{X_n}}(x_0):=\frac{\sum_{x_n}\lambda_{x_n} P(x_0,x_n) }{\sum_{x_n} \lambda_{x_n} P(x_n)}\,,
\end{equation}

where the probabilities in the right-hand side are obtained by marginalization of the joint PMF of the BN. Eq.~\eqref{eq:updatingVE} can be equivalently obtained by augmenting the BN with auxiliary binary node $D_{X_n}$ as a child of $X_n$. By specifying $P(d_{X_n}|x_n):=\lambda_{x_n}$ for each $x_n \in \Omega_{X_n}$, it is easy to check that $P(x_0|d_{X_n})=P_{\lambda_{X_n}}(x_0)$, i.e. Eq.~\eqref{eq:updatingVE} can be reduced to a standard updating in an augmented BN.

The notion of \emph{soft evidence} (SE) refers to a different situation, in which the observational process returns an elicitation $P'(X_n)$ for the marginal PMF of $X_n$. See \cite{mrad2015explication} for a detailed discussion on the possible situations producing SE. If this is the case, $P'(X_n)$ is assumed to replace the original beliefs about $X_n$ by \emph{Jeffrey's updating} \cite{jeffrey1965}, i.e.
\begin{equation}\label{eq:updatingSE}
P_{X_n}'(x_0) := \sum_{x_n} P(x_0|x_n) \cdot P'(x_n)\,.
\end{equation}

Eq.~\eqref{eq:updatingSE} for SE reduces to Eq.~\eqref{eq:updating} whenever $P'(X_n)$ assigns all the probability mass to a single value in $\Omega_{X_n}$. The same happens for VE in Eq.~\eqref{eq:updatingVE}, when all the likelihoods are zero apart from the one corresponding to the observed value. Although SE and VE refer to epistemologically different informational settings, the following result provides means for a unified approach to their modeling.

\begin{proposition}[\cite{chan2005}]\label{prop:SEVEequiv}
Absorption of a SE $P'(X_n)$ as in Eq.~\eqref{eq:updatingSE} is equivalent to Eq.~\eqref{eq:updatingVE} with a VE specified as:

\begin{equation}\label{eq:l2P}
\lambda_{x_n} \propto \frac{P'(x_n)}{P(x_n)}\,,
\end{equation}
for each $x_n \in \Omega_{X_n}$.\footnote{VE is defined as a collection of likelihoods, which in turn are defined up to a multiplicative positive constant. This clearly follows from Eq.~\eqref{eq:updatingVE}. The relation in Eq.~\eqref{eq:l2P} is proportionality and not equality just to make all the likelihoods smaller or equal than one.}

Vice versa, absorption of a VE $\lambda_{X_n}$ as in Eq.~\eqref{eq:updatingVE} is equivalent to Eq.~\eqref{eq:updatingSE} with a SE specified as:

\begin{equation}\label{eq:P2l}
P'(x_n) := \frac{\lambda_{x_n} P(x_n)}{\sum_{x_n} \lambda_{x_n} P(x_n)}\,,
\end{equation}
for each $x_n \in \Omega_{X_n}$.
\end{proposition}

In the above setup for SE, states that are impossible in the original BN cannot be revised, i.e. if $P(x_n)=0$ for some $x_n\in\Omega_{X_n}$, then also $P'(x_n)=0$ and any value can be set for $\lambda_{x_n}$. \emph{Vice versa}, according to Eq.~\eqref{eq:P2l}, a zero likelihood in a VE renders impossible the corresponding state of the SE. Thus, at least a non-zero likelihood should be specified in a VE. All these issues are shown in the following example.

\begin{example}\label{ex:vese}
Let $X$ denote the actual color of a traffic light with $\Omega_X:=\{g,y,r\}$. Assume $g$ (green) more probable than $r$ (red), and $y$ (yellow) impossible. Thus, for instance, $P(X)=[\sfrac{4}{5},0,\sfrac{1}{5}]$. We eventually revise $P(X)$ by a SE $P'(X)$, which keeps yellow impossible and assigns the same probability to the two other states, i.e. $P'(X)=[\sfrac{1}{2},0,\sfrac{1}{2}]$. Because of Eq.~\eqref{eq:l2P}, this can be equivalently achieved by a VE $\lambda_X \propto \{ 1, 1, 4 \}$. Vice versa, because of Eq.~\eqref{eq:P2l}, a VE $\tilde{\lambda}_X \propto \{ 1, 1, 5\}$ induces an updated $P_{\tilde{\lambda}}(X)=[\sfrac{4}{9},0,\sfrac{5}{9}]$. Such PMF coincides with $P(X|d_X)$ in a two-node BN, with $D_X$ child of $X$, CPT $P(D_X|X)$ with $P(d_X|X)=[\sfrac{1}{10},\sfrac{1}{10},\sfrac{1}{2}]$ and marginal PMF $P(X)$ as in the original specification.
\end{example}

\section{Credal Uncertain Evidence}
\subsection{Credal Virtual Evidence}
We propose \emph{credal} VE (CVE) as a robust extension of sharp virtual observations. Notation ${\Lambda}_{X_n}$ is used here for the intervals $\{ \underline{\lambda}_{x_n},\overline{\lambda}_{x_n}\}_{x_n\in\Omega_{{X}_n}}$. CVE updating is defined as the computation of the bounds of Eq.~\eqref{eq:updatingVE} with respect to all VEs $\lambda_{X_n}$ consistent with the interval constraints in $\Lambda_{X_n}$. Notation $\underline{P}_{\Lambda_{X_n}}(x_0)$ and $\overline{P}_{\Lambda_{X_n}}(x_0)$ is used to denote these bounds. CVE absorption in BNs is done as follows.

\begin{transformation}\label{tr:cve}
Given a BN over $\bm{X}$ and a CVE $\Lambda_{X_n}$, add a binary child $D_{X_n}$ of $X_n$ and quantify its CCPT $K(D_{X_n}|{X_n})$ with constraints $\underline{\lambda}_{x_n} \leq P(d_{X_n}|x_n) \leq \overline{\lambda}_{x_n}$.\footnote{For binary $B$, constraint $l\leq P(b)\leq u$ defines a CS $K(B)$ with elements $P_1(B):=[l,1-l]$ and $P_2(B):=[u,1-u]$.} A CN with a single credal node results.
\end{transformation}

By Tr.~\ref{tr:cve}, CVE updating in a BN is reduced to CN updating.

\begin{theorem}\label{th:cve}
Given a CVE in a BN, consider the CN returned by Tr.~\ref{tr:cve}. Then:
\begin{equation}
\underline{P}(x_0|d_{X_n})=\underline{P}_{\Lambda_{X_n}}(x_0)\,,
\end{equation}
and analogously for the upper bounds.
\end{theorem}

Standard VE can be used to model partially reliable sensors or tests, whose quantification is based on sensitivity and specificity data. Since these data are not always promptly/easily available (e.g., a pregnancy test whose failure can be only decided later), a CVE with interval likelihoods can be quantified by the \emph{imprecise Dirichlet model}\footnote{Given $N$ observations of $X$, if $n(x)$ of them reports $x$, the lower bound of $P(x)$ for to the imprecise Dirichlet model is $\frac{n(x)}{N+s}$, and the upper bound $\frac{n(x)+s}{N+s}$, with $s$ effective prior sample size.} \cite{bernardIDM} as in the following example.

\begin{example}\label{ex:acl}
The reference standard for diagnosis of anterior cruciate legament sprains is arthroscopy. In a trial, 40 patients coming in with acute knee pain are examined using the \emph{Declan test} \cite{declan}. Every patient also has an arthroscopy procedure for a definitive diagnosis. Results are TP=17 (Declan positive, arthroscopy positive), FP=3 (Declan positive, arthroscopy negative), FN=6 (Declan negative, arthroscopy positive) and TN=14 (Declan negative, arthroscopy negative). Patients visiting a clinic have prior sprain probability $P(x)=0.2$. Given a positive Declan, the imprecise Dirichlet model (see Footnote 3) with $s=1$ corresponds to CVE $\underline{\lambda}_{x}=\sfrac{17}{23+1}$, $\overline{\lambda}_{x}=\sfrac{17+1}{23+1}$, $\underline{\lambda}_{\neg x}=\sfrac{3}{17+1}$, $\overline{\lambda}_{\neg x}=\sfrac{3+1}{17+1}$. The bounds of the updated sprain probability with respect to the above constraints are $\underline{P}_{\Lambda_X}(x)=\sfrac{1}{3}$, $\overline{P}_{\Lambda_X}(x)\simeq 0.53$. A VE with frequentist estimates would have produced instead $P_{\lambda_X}\simeq 0.51$.
\end{example}

\subsection{Credal Soft Evidence}
Analogous to CVE, \emph{credal soft evidence} (CSE) on $X_n$ can be specified by any CS $K'(X_n)$. Accordingly, CSE updating computes the bounds spanned by the updating of all SEs based on PMFs consistent with the CS, i.e.
\begin{equation}\label{eq:credalJeffrey}
\underline{P}'_{X_n}(x_0):=\min_{P'(X_n)\in K'(X_n)} \sum_{x_n} P(x_0|x_n) \cdot P'(x_n)\,,
\end{equation}
and analogously for the upper bound $\overline{P}'_{X_n}(x_0)$.

The \emph{shadow} of a CS $K(X)$ is a CS $\hat{K}(X)$ obtained from all the PMFs $\hat{P}(X)$ such that, for each $x\in\Omega_X$:
\begin{equation}
\min_{P(X)\in K(X)} P(x) \leq \hat{P}(x) \leq \max_{P(X)\in {K}(X)} P(x)\,.
\end{equation}
A CS coinciding with its shadow is called \emph{shady}. It is a trivial exercise to check that CSs over binary variables are shady.
\footnote{Following \cite{campos1994}, a shadow is just the set of probability intervals induced by a generic CS.}

The following result extends Pr.~\ref{prop:SEVEequiv} to the imprecise framework.

\begin{theorem}\label{th:CSECVEequiv}
Absorption of a CSE with shady $K'(X_n)$ is equivalent to that of CVE $\Lambda_{X_n}$ such that:
\begin{equation}\label{eq:Th2a}
\underline{\lambda}_{x_n} \propto \frac{\underline{P}'(x_n)}{P(x_n)}\,,
\end{equation}
where $\underline{P}'(x_n):=\min_{P'(X_n)\in K'(X_n)} P'(x_n)$ and analogously for the upper bound. Vice versa absorption of a CVE $\Lambda_{X_n}$ is equivalent to that of a CSE such that:
\begin{equation}\label{eq:Th2b}
\underline{P}'(x_n) = \frac{P(x_n)\underline{\lambda}_{x_n}} {P(x_n)\underline{\lambda}_{x_n}+\sum_{x_n'\neq x_n} P(x_n') \overline{\lambda}_{x_n'}}\,,
\end{equation}
and analogously with a swap between lower and upper likelihoods for the upper bound.
\end{theorem}

By Th.~\ref{th:cve} and ~\ref{th:CSECVEequiv}, CSE updating in a BN is reduced to standard updating in a CN. This represents a generalization to the credal case of Pr.~\ref{prop:SEVEequiv}. For CSEs with non-shady CSs, the procedure is slightly more involved, as detailed by the following result.

\begin{proposition}\label{pr:nonshady}
Given a CSE $K'(X_n):=\{ P_i'(X_n)\}_{i=1}^k$ in a BN, add a binary child $D_{X_n}$ of $X_n$ quantified by an ECPT $\{P_i(D_{X_n}|X_n) \}_{i=1}^k$ such that $P_i(d_{X_n}|x_n) \propto \frac{P_i'(x_n)}{P(x_n)}$ for each $i=1,\ldots,k$ and $x_n \in \Omega_{X_n}$. Then:
\begin{equation}
\underline{P}'_{X_n}(x_0)=\underline{P}(x_0|d_{X_n})\,.
\end{equation}
\end{proposition}

To clarify these results, consider the following example.

\begin{example}
Consider the same setup as in Ex.~\ref{ex:vese}. Let us revise the original PMF $P(X)$ by a CSE based on the shady CS $K'(X):=\{ P_1'(X),P_2'(X)\}$, with $P_1'(X):=[0.6,0,0.4]$ and $P_2'(X):=[0.4,0,0.6]$. Th.~\ref{th:CSECVEequiv} can be used to convert such CSE in a CVE $\Lambda_X:=\{2\textnormal{-}3:1:8\textnormal{-}12\}$. Vice versa, the beliefs induced by CVE $\tilde{\Lambda}_X:=\{3\textnormal{-}5:1:8\textnormal{-}10\}$ are $\underline{P}_{\tilde{\Lambda}_X}(g)=\sfrac{3}{5}$, $\overline{P}_{\tilde{\Lambda}_X}(g)=\sfrac{2}{3}$, $\underline{P}_{\tilde{\Lambda}_X}(y)=\underline{P}_{\tilde{\Lambda}_X}(y)=0$, and $\underline{P}_{\tilde{\Lambda}_X}(r)=\sfrac{1}{3}$, $\overline{P}_{\tilde{\Lambda}_X}(r)=\sfrac{2}{5}$. These bounds may be equivalently obtained in a two-node CN with $D_X$ child of $X$ and CCPT $K(D_X|X)$ such that $P(d_X|X=g)\in[0.6,1]$, $P(d_X|X=y)=1$, and $P(d_X|X=r)\in[0.8,1]$. Alternatively, following Pr.~\ref{pr:nonshady}, absorption of $K'(X)$ can be achieved by a ECCPT with two CPTs.
\end{example}

We point out that \emph{conservative updating} (CU), a credal updating rule for reliable treatment of missing non-MAR data \cite{decooman2004}, falls as a special case in our formalism. CU is defined as:
\begin{equation}\label{eq:cu}
\underline{P}_{X_n}'(x_0)=\min_{x_n\in\Omega_{X_n}} P(x_0|x_n)\,,
\end{equation}
and represents the most conservative approach to belief revision. A \emph{vacuous} CCPT is specified, with $[0,1]$ intervals for each value, either i) by Tr.~\ref{tr:cve}, given CVE whose likelihoods take any value between zero and one \footnote{As VE likelihoods are defined up to a positive multiplicative constant, we can set any positive $\overline{\lambda}_{x_n}$ provided that $\underline{\lambda}_{x_n}=0$.}, or ii) by straightforward application of Th.~\ref{th:CSECVEequiv}, if a \emph{vacuous} CSE $K_0'(X_n)$ is provided. The resulting ECPT with $|\Omega_{X}|$ CPTs\footnote{The induced ECPT contains all $2^{|\Omega_{X_n}|}$ combinations of zero and ones in the CPTs. Yet, only those having a single one in the row associated to $d_{X_n}$ remains after the convex hull.} corresponds to the CU implementation in \cite{antonucci2008}.  Also, Eq.~\eqref{eq:credalJeffrey} reduces to Eq.~\eqref{eq:cu}, given vacuous CSE. We can similarly proceed in the case of \emph{incomplete} observations, i.e. some values of $X_n$ are recognized as impossible, but no information can be provided about the other ones. If this is the case, we just replace $\Omega_{X_n}$ with $\Omega_{X_n}'\subset\Omega_{X_n}$.
	
\section{Credal Probability Kinematics}
Given two joint PMFs $P(\bm{X})$ and $P'(\bm{X})$, we say that the latter comes from the first by \emph{probability kinematics} (PK) on the (coarse) partition of $\Omega_{\bm{X}}$ induced by $X_n$ if and only if $P'(\bm{x}|x_n)=P(\bm{x}|x_n)$ for each $\bm{x}\in\Omega_{\bm{X}}$ and $x_n\in\Omega_{X_n}$ \cite{diaconis1982,chan2005}.\footnote{Full consistency of $P'$ with the evidence inducing the revision process is not explicitly required. A more stringent characterization of PK was proposed, among others, by \cite{wagner2002probability}} This is the underlying assumption in Eq.~\eqref{eq:updatingSE}. If $P'(\bm{X})$ is replaced by a CS, PK is generalized as follows.

\begin{definition}
Let $P(\bm{X})$ and $K'(\bm{X})$ be, respectively, a joint PMF and a joint CS. We say that $K'(\bm{X})$ comes from $P(\bm{X})$ by \emph{credal probability kinematics} (CPK) on the partition of $\Omega_{\bm{X}}$ induced by $X_n$ if and only if it holds $\underline{P}'(\bm{x}|x_n)={\overline{P}}'(\bm{x}|x_n)=P(\bm{x}|x_n)$, for each $\bm{x}\in\Omega_{\bm{X}}$ and $x_n\in\Omega_{X_n}$.
\end{definition}

That is, any revision process based on (generalized) PK guarantees invariance of the relevance of $x_n$, for each $x_n\in\Omega_{X_n}$, to any other possible event in the model, say $x_0$. The following consistency result holds for CSEs.

\begin{theorem} 
Given a BN over $\bm{X}$ and a shady CSE $K'(X_n)$, convert the CSE into a CVE as in Th.~\ref{th:CSECVEequiv} and transform the BN into a CN by Tr.~\ref{tr:cve}. Let $K'(\bm{X},D_{X_n})$ be the joint CS associated to the CN. Then, $K'(\bm{X}|d_{X_n})$ comes from $P(\bm{X})$ by CPK on the partition induced by $X_n$. Moreover $K'(X_n|d_{X_n})$ coincides with the marginal CS in the CN.
\end{theorem}

\section{Multiple Evidences}
So far, we only considered the updating of a \emph{single} CVE or CSE. We call \emph{uncertain credal updating} (UCU) of a BN the general task of computing updated/revised beliefs in a BN with an arbitrary number of CSEs, CVEs, and hard evidences as well. Here, UCU is intended as iterated application of the procedures outlined above. See for instance \cite{dubois2008three}, for a categorization of iterated belief revision problems and their assumptions. When coping with multiple VEs in a BN, it is sufficient to add the necessary auxiliary children to the observed variables and quantify the CPTs as described. We similarly proceed with multiple CVEs.

The procedure becomes less straightforward when coping with multiple SEs or CSEs, since quantification of each auxiliary child by Eq.~\eqref{eq:l2P} requires a preliminary inference step. As a consequence, iterated revision might be not invariant with respect to the revision process scheme \cite{wagner2002probability}. 

Additionally, with CSEs, absorption of the first CSE transforms the BN into a CN, and successive absorption of other CSEs requires further extension of the procedure in Th.~\ref{th:CSECVEequiv}. We leave such an extension as future work, and here we just consider simultaneous absorption of all evidences. If this is the case, multiple CSEs can be converted in CVEs and the inferences required for the quantification of the auxiliary children is performed in the original BN. 

\subsection{Algorithmic and Complexity Issues}
\emph{ApproxLP} \cite{antonucci2014e} is an algorithm for general CN updating based on linear programming. It provides an inner approximation of the updated intervals with the same complexity of a BN inference on the same graph. Roughly, CN updating is reduced by ApproxLP to a sequence of linear programming tasks. Each is obtained by iteratively fixing all the local models to single elements of the corresponding CSs, while leaving a \emph{free} single variable. It follows the algorithm efficiently produces exact inferences whenever a CN has all local CSs made of a single element apart from one. This is the case of belief updating with a single CVE/CSE.

\subsection{Complexity Issues}
Since standard BN updating of polytrees can be performed efficiently, the same happens with VEs and/or SEs, as Tr.~\ref{tr:cve} does not affect the topology (nor the treewidth) of the original network. Similarly, with multiply connected models, BN updating is exponential in the treewidth, and the same happens with models augmented by VEs and/or SEs.

As already noticed, with CNs, binary polytrees can be updated efficiently, while updating ternary polytrees is already NP-hard. An important question is therefore whether or not a similar situation holds for UCU in BNs. The (positive) answer is provided by the two following results.

\begin{proposition}
UCU of polytree-shaped binary BNs can be solved in polynomial time.
\end{proposition}

The proof of this proposition is trivial and simply follows from the fact that the auxiliary nodes required to model CVE and/or CSE are binary (remember that CSs over binary variables are always shady). The CN solving the UCU is therefore a binary polytree that can be updated by the exact algorithm proposed in \cite{fagiuoli1998}. 

\begin{theorem}\label{th:complexity}
UCU of non-binary polytree-shaped BNs is NP-hard.
\end{theorem}

The proof of this theorem is based on a reduction to the analogous result for CNs \cite{maua2014a}. This already concerns models whose variables have no more than three states and treewidth equal to two. In these cases, approximate inferences can be efficiently computed by ApproxLP.

\section{Credal Opinion Pooling}
Consider the generalized case of $m\geq 1$ overlapping probabilistic instances on $X_n$. For each $j=1,\ldots,m$, let $P'_j(X_n)$ denote the SE reported by the $j$-th source. Straightforward introduction of $m$ auxiliary nodes as outlined above would suffer \emph{confirmational} dynamics, analogous to the well-known issue with posterior probability estimates in the naive Bayes classifier \cite{rish2001}. This might likely yield inconsistent revised beliefs, i.e. $\tilde{P}'(X_n)$ falls outside the convex hull of $\{P_j'(X_n)\}_{j=1}^m$.

A most conservative approach to prevent such inconsistency adopts the convex hull of all the opinions \cite{stewart2017}. In our formalism, this is just the CS $K'(X_n) := \{P_j'(X_n)\}_{j=1}^m$. Yet, consider any small $\epsilon>0$, and assume $P_1'(x_n)=\epsilon$, $P_2'(x_n)=1-\epsilon$, and $P_j'(x_n)=p \in (\epsilon,1-\epsilon)$ for each $j=3,\ldots,m$. Despite the consensus of all remaining sources on sharp value $p$, the conservative approach above would yield $K'(X_n)\simeq K_0(X_n)$. To what extent should this be preferred to the \emph{confirmational} case is an open question.

A compromise solution might be offered by the \emph{geometric pooling operator} (or LogOp) \cite{bacharach1975}. Given a collection of positive weights $\{ \alpha_j \}_{j=1}^m$, with $\sum_{j=1}^m \alpha_j = 1$, the LogOp functional produces the PMF $\tilde{P}'(X_n)$ such that:
\begin{equation}\label{eq:logop}
\tilde{P}'(x_n) \propto \prod_{j=1}^m P_j'(x)^{\alpha_j}\,,
\end{equation}
for each $x_n\in\Omega_{X_n}$. $\tilde{P}'(X_n)$ belongs to the convex hull of $\{P_j'(X_n)\}_{j=1}^m$ for any specification of the weights \cite{adamvcik2014information}. The overlapping SEs associated to the PMF in Eq.~\eqref{eq:logop} can be equivalently modeled by a collection of $m$ VEs defined as follows.

\begin{transformation}\label{tr:pool}
Consider a BN over $\bm{X}$ and a collection of SEs on $X_n$, $\{ P_j'(X_n) \}_{j=1}^m$. For each $j=1,\ldots,m$, augment the BN with binary child $D_{X_n}^{(j)}$ of $X_n$ whose CPT is such that $P(d_{X_n}^{(j)}|x_n)\propto\left[\frac{P'(x_n)}{P(x_n)}\right]^{\alpha_j}$, with $\sum_{j=1}^m\alpha_j=1$.
\end{transformation}

The transformation is used for the following result.

\begin{proposition}\label{pr:logop}
Consider the same inputs as in Tr.~\ref{tr:pool}. Then:
\begin{equation}\label{eq:pooling_th}
\tilde{P}_{X_n}'(x_0) = P(x_0|d_{X_n}^{(1)},\ldots,d_{X_n}^{(m)})\,,
\end{equation}
where the probability on the left-hand side is obtained by the direct revision induced by $\tilde{P}'(X_n)$, while the probability on the right-hand side of Eq.~\eqref{eq:pooling_th} has been computed in the BN returned by Tr.~\ref{tr:pool}.
\end{proposition}

The proof follows from the conditional independence of the auxiliary nodes given $X_n$. Also, note how our proposal simultaneously performs pooling and absorption of overlapping SEs.

Suppose $m$ sources provide generalized CSEs about $X_n$, say $\{K_j'(X_n)\}_{j=1}^m$.
Let $\tilde{K}'(X_n)$ denote the CS induced by LogOp as in Eq.~\eqref{eq:logop}, for each $P_j'(X_j)\in K_j'(X_n)$, $j=1,\ldots,m$ \cite{adamvcik2014information}. We generalize Tr.~\ref{tr:pool} as follows:
\begin{transformation}\label{tr:pool_credal}
Consider a BN over $\bm{X}$ and the collection of CSEs $\{ K_j'(X_n)\}_{j=1}^m$. For each $j=1,\ldots,m$, augment the BN with binary child $D_{X_n}^{(j)}$ of $X_n$, whose CCPT is such that $\underline{P}(d_{X_n}^{(j)}|x_n)\propto\left[\frac{\underline{P}'(x_n)}{P(x_n)}\right]^{\alpha_j}$ and $\overline{P}(d_{X_n}^{(j)}|x_n)\propto\left[\frac{\overline{P}'(x_n)}{P(x_n)}\right]^{\alpha_j}$.
\end{transformation}

This transformation returns a CN. A result analogous to Pr.~\ref{pr:logop} can be derived.

\begin{theorem}
Consider the same inputs as in Tr. \ref{tr:pool_credal}. Then:
\begin{equation}\label{eq:clogopeq}
\underline{\tilde{P}}_{X_n}'(x_0)=\underline{P}(x_0|d_{X_n}^{(1)},\dots,d_{X_n}^{(m)})\,,
\end{equation}
where the lower probability on the left-hand side has been computed by absorption of the single CSE $\tilde{K}'(X_n)$ and the probability on the right-hand side has been computed in the CN returned by Tr.~\ref{tr:pool_credal}. The same relation also holds for the corresponding upper probabilities.
\end{theorem}

\section{Conclusions}
Credal, or set-valued, modeling of uncertain evidence has been proposed within the framework of Bayesian networks. Such procedure generalizes standard updating. More importantly, our proposal allows to reduce the task of absorption of uncertain evidence to standard updating in credal networks. Complexity results, specific inference schemes, and generalized pooling procedures have been also derived. 

As a future work we intend to evaluate the proposed technique with knowledge-based decision-support systems based on Bayesian network to model unreliable observational processes. Moreover the proposed procedure should be extended to the framework of credal networks, thus reconciling the orthogonal viewpoints considered in this paper and in \cite{decampos}, and tackling the case of non-simultaneous updating.

\appendix
\section{Proofs}

\begin{myproof}
The proof follows from the analogous result with BNs. For any BN consistent with the CN returned by Tr.~\ref{tr:cve}, we have:
\begin{align*}
P(x_0|d_{X_n}) &= \frac{P(x_0,d_{X_n})}{P(d_{X_n})}\\
&=\frac{\sum_{x_n} P(x_0|x_n)P(d_{X_n}|x_n)P(x_n)}{\sum_{x_n} P(d_{X_n}|x_n) P(x_n)}\,.
\end{align*}
As $P(d_{X_n}|x_n)$ reaches its minimum at $\underline{\lambda}_{x_n}$, for every $x_n \in \Omega_{X_n}$, the minimization of the last term coincides with that of Eq.~\eqref{eq:updatingVE} and gives $\underline{P}_{\Lambda_{X_n}}(x_0)$, the other elements being constant. Analogous reasoning yields $\overline{P}_{\Lambda_{X_n}}(x_0)$.\\
\qed
\end{myproof}

For the proof of Th.~\ref{th:CSECVEequiv}, we need to introduce the following transformation and lemma.
\begin{transformation}\label{tr:cveext}
Consider a CSE $K'(X_n)$ in a BN. Let $\{P_i'(X_n), i=1\dots,n_v\}$ denote the elements of the CS $K'(X_n)$.\footnote{Remember that in our definition of CS we remove the inner points of the convex hull.} In the BN, compute the marginal PMF $P(X_n)$ with standard algorithms. Augment the BN with a binary node $D_{X_n}$, such that $\Pi_{X_n} := \{X_n\}$. Quantify the local model for $D_{X_n}$ as an ECPT $K(D_{X_n}|X_n)$, specified as a set of $n_v$ CPTs $\{P_i(D_{X_n}|X_n): i=1,\dots,n_v\}$. $P_i(D_{X_n}|X_n)$ is defined as:
\begin{equation}\label{eq:proport}
P_i(d_{X_n}|x_n) \propto \frac{P_i'(x_n)}{P(x_n)}\,,
\end{equation}
for each $i=1,\ldots,n_v$.
The same prescriptions provided after Pr.~\ref{prop:SEVEequiv} for the case of zero-probability events should be followed here.
\end{transformation}

\begin{lemma}\label{lm:cve}
Given a CSE $K'(X_n)$ in a BN, consider the CN returned by Tr.~\ref{tr:cveext}. Then:
\begin{equation}
\underline{P}(x_0|d_{X_n})=\underline{P}_{X_n}'(x_0)\,,
\end{equation}
and analogously for the upper bound.
\begin{proof}
$D_{X_n}$ is the only credal node in the CN. Thus:
\begin{equation}\label{eq:lm1}
\underline{P}(x_0|d_{X_n}) = \min_{P(d_{X_n}|X_n)\in K(d_{X_n}|X_n)} \frac{P(x_0,d_{X_n})}{P(d_{X_n})}\,.
\end{equation}
Let us rewrite Eq.~\eqref{eq:lm1} by: (i) explicitly enumerating the CPTs in the ECPT $K(D_{X_n}|X_n)$, (ii) making explicit the marginalization of $X_n$, (iii) exploiting the fact that, by the Markov condition, we have conditional independence between $D_{X_n}$ and $X_0$ given $X_n$. The result is:
\begin{equation}
\min_{i=1,\ldots,n_v} \frac{\sum_{x_n} P(x_0|x_n) \cdot P_i(d|x_n) \cdot P(x_n)}{\sum_{x_n} P_i(d|x_n)\cdot P(x_n)}\,.
\end{equation}
Thus, because of Eq.~\eqref{eq:proport}:
\begin{equation}\label{eq:lm2}
\min_{i=1,\ldots,v} \frac{\sum_{x_n} P(x_0|x_n) \cdot P_i'(x_n)}{\sum_{x_n} P_i'(x_n)}\,.
\end{equation}
As the denominator in Eq.~\eqref{eq:lm2} is one we obtain Eq.~\eqref{eq:credalJeffrey}. This proves the lemma.\\
\end{proof}
\end{lemma}
We can now prove the second theorem.
\begin{myproof}
Let us first prove the second part of the theorem. As a consequence of Pr.~\ref{prop:SEVEequiv}, each VE consistent with the CVE can be converted in a SE defined as in Eq.~\eqref{eq:P2l}. The CS implementing the CSE equivalent to the CVE is therefore:
\begin{equation}
K'(X_n):=
\left\{
P'(X_n) \left| 
\substack{P'(x_n)=\frac{P(x_n) \lambda_{x_n}}{\sum_{x_n} P(x_n) \lambda_{x_n}} \\
	\underline{\lambda}_{x_n} \leq \lambda_{x_n} \leq \overline{\lambda}_{x_n} \forall x_n}
\right.
\right\}\,.
\end{equation}
The computation of $\underline{P}'(x_n)$ is therefore a linearly constrained linear fractional task. If $P(x_n)>0$, we can rewrite the objective function as:
\begin{equation}
P'(x_n)= \left[ 1+\sum_{x_n'\neq x_n} \frac{\lambda_{x_n'}P(x_n')}{\lambda_{x_n}P(x_n)} \right]^{-1} \,.
\end{equation}
As $f(\alpha)=(1+\alpha)^{-1}$ is a monotone decreasing function of $\alpha$, minimizing the objective function is equivalent to maximize:
\begin{equation}\label{eq:obj}
\sum_{x_n'\neq x_n}\frac{\lambda_{x_n'}P(x_n')}{\lambda_{x_n}P(x_n)}\,,
\end{equation}
and vice versa for the maximization. As each $\lambda_{x_n}$ can vary in its interval independently of the others, the maximum of the function in Eq.~\eqref{eq:obj} is obtained by maximizing the numerator and minimizing the denominator, i.e., for $\lambda_{x_n}=\overline{\lambda}_{x_n}$ and $\lambda_{x_n'}=\underline{\lambda}_{x_n'}$. This proves Eq.~\eqref{eq:Th2b}, which remains valid also for $P(x_n)=0$.

To prove the first part of the theorem, because of Lm.~\ref{lm:cve}, we only need to prove that the CN returned by Tr.~\ref{tr:cveext} and the CN returned by Tr.~\ref{tr:cve} for the CVE specified in Eq.~\eqref{eq:Th2a} provides the same $\underline{P}(x_0|x_n)$. This lower posterior probability in the second CN rewrites as:

\begin{equation}\label{eq:flp}
\underline{P}(x_0|d_{X_n})=\!\!\!\min_{\underline{\lambda}_{x_n} \leq \lambda_{x_n} \leq \overline{\lambda}_{x_n}} \frac{\sum_{x_n} P(x_0|x_n) \lambda_{x_n} P(x_n)}{\sum_{x_n} \lambda_{x_n} P(x_n)}\,.
\end{equation}
	
Again, this is a linearly constrained linear fractional task, which can be reduced to a linear task by  \cite{boyd2004convex}. In the linear task, the minimum is achieved when the $\lambda_{x_n}$ corresponding to the maximum \emph{coefficient} $P(x_0|x_n) P(x_n)$ of the numerator of the objective function takes the minimum value $\underline{\lambda}_{x_n}$. But as $\underline{\lambda}_{x_n} = \min_i \frac{P_i'(x_n)}{P(x_n)}$, we can equivalently obtain this value with the ECPT in the first CN. This proves the first part of the theorem.\\
\qed
\end{myproof}

\begin{myproof} 
The result follows from the analogous for PK. Thus, let us first assume $K'(X_n)$ composed of a single PMF $P'(X_n)$. This means that the CSE degenerates into a standard SE. Let $\lambda_{X_n}$ denote the corresponding VE and consider the augmented BN obtained by adding the auxiliary binary child $D_{X_n}$. Also, let $\bm{x} \in \Omega_{X_n}$ be any configuration of the joint variable $\bm{X}$. By the Markov condition:
\begin{align*}
P'(\bm{x}|x_n)&= P(\bm{x}|x_n,d_{X_n})\\
&= \frac{P(\bm{x}|x_n)P(x_n)P(d_{X_n}|x_n)}{P(x_n)P(d_{X_n}|x_n)}\\
&= P(\bm{x}|x_n)\,.
\end{align*}
For a CSE $K'(X_n)$ including more than a PMF, we just repeat the same above considerations separately for each $P'(X_n) \in K'(X_n)$ and obtain the proof of the statement. Also, by Th.~(2) it holds $K'(\bm{x})^{\downarrow X_n} = K'(x_n)$, for all configurations $\bm{x}$ consistent with $X_n = x_n$, for all $x_n \in X_n$.\\
\qed
\end{myproof}

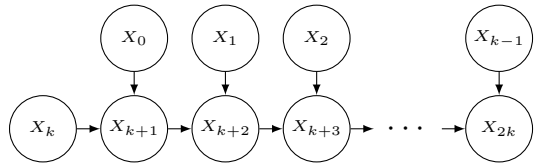
\begin{figure}[htp!]
	\centering
	\begin{tikzpicture}
	\tikzstyle{every node}=[circle,draw,minimum height=25pt,inner sep=1pt]
	\tikzstyle{every path}=[->,>=latex]
	
	\node (X0) at (-1.2,0) {\tiny $X_{k}$};
	\node (X1) at (0,1.2) {\tiny $X_0$};
	\node (X3) at (1.2,1.2) {\tiny $X_1$};
	\node (X5) at (2.4,1.2) {\tiny $X_2$};
	\node (X5b) at (4.8,1.2) {\tiny $X_{k-1}$};
	
	\node (X2) at (0,0) {\tiny $X_{k+1}$};
	\node (X4) at (1.2,0) {\tiny $X_{k+2}$};
	\node (X6) at (2.4,0) {\tiny $X_{k+3}$};
	\node[draw=none] (ret) at (3.6,0) {$\dotsb$};
	\node[] (X6b) at (4.8,0) {\tiny $X_{2k}$};
	\draw (X0) -- (X2);   
	\draw (X1) -- (X2);   
	\draw (X3) -- (X4);   
	\draw (X2) -- (X4);   
	\draw (X4) -- (X6);
	\draw (X5) -- (X6);
	\draw (X5b) -- (X6b);
	\draw (X6) -- (ret);
	\draw (ret) -- (X6b);
	\end{tikzpicture}
\caption{A polytree-shaped directed acyclic graph.}
\label{fig:denis}
\end{figure}

\begin{myproof} 
To prove the theorem we show that the non-binary polytree-shaped CN used by \cite[Th.~1]{maua2014a} to prove the NP-hardness of non-binary credal polytrees can be used to model UCU in a non-binary polytree-shaped BN. To do that for an arbitrary $k$, consider the BN over $\bm{X}:=(X_0,X_1,\ldots,X_{2k})$ with the topology in Fig.~\ref{fig:denis}, Nodes $(X_0,\ldots,X_{k=1})$ are associated to binary variables, the others to ternary variables. A uniform marginal PMF is specified for $X_k$, while the CPTs for the other ternary variables are as indicated in Table~2 of the proof we refer to (the numerical values being irrelevant for the present proof). For the binary variables we also specify a uniform prior. 

We specify indeed a \emph{vacuous} CSE for each binary variable. These CSEs can be asborbed by replacing the uniform PMFs with vacuous CSs. The resulting model is exactly the CN used to reduce CN updating to the \emph{PARTITION} problem \cite{garey:1979} and hence proves the thesis.\\
\qed
\end{myproof}

\begin{myproof}
For any BN consistent with the CN resulting from Tr.~\ref{tr:pool_credal} it holds $P_{LogOp_{\bm{\alpha},\bm{P}'}}(x_0) = P(x_0 | d_{X_n}^{(1)},\dots,d_{X_n}^{(m)})$, by Prop.~\ref{pr:logop}. By definition, see Eq.~\eqref{eq:logop}, we have:

\begin{equation}\label{eq:minclogop}
\min_{P_j \in K_j, K_j \in \bm{K}'} cLogOp_{\bm{K'}}^{\bm{\alpha}}(x_n) = k \prod_{j=1}^m \underline{P}'_j(x_n)^{\alpha_j} \,,
\end{equation}
with $k$ being the normalization constant and $\underline{P}'_j(x_n) = \min_{P \in K'_j(x_n)} P(x_n)$, for every $j=1,\dots,m$ and for all $x_n \in \Omega_{X_n}$.

It follows:
\begin{align*}
\min_{P(x_n) \in cLogOp_{\bm{K}'}^{\bm{\alpha}}(x_n)} \tilde{P}(x_0) &= k \sum_{x_n} P(x_0|x_n) \prod_{j=1}^n \underline{P}'_j(x_n)\\
&= \underline{P}(x_0| d_{X_n}^{(1)},\dots,d_{X_n}^{(m)})\,,
\end{align*}
where the second term comes by Eq.~\eqref{th:CSECVEequiv} and Eq.~\eqref{eq:minclogop}. This gives the proof of the theorem.
\end{myproof}

\bibliographystyle{amsplain} 
\bibliography{bibfile.bib}

\end{document}